\pdfoutput=1
\documentclass[11pt,a4paper]{article}
\usepackage[hyperref]{emnlp2020}
\usepackage{times}
\usepackage{latexsym}
\usepackage{tabularx}
\usepackage{graphicx}
\usepackage{adjustbox}
\usepackage{multirow}
\usepackage{booktabs}
\usepackage{amsmath}
\usepackage{todonotes}
\usepackage{float}
\restylefloat{table}
\usepackage{hyperref}

\definecolor{cadmiumorange}{rgb}{0.93, 0.53, 0.18}
\definecolor{darkspringgreen}{rgb}{0.09, 0.45, 0.27}
\definecolor{chamoisee}{rgb}{0.63, 0.47, 0.35}

\usepackage{microtype}

\usepackage{subfiles}

\aclfinalcopy 

\title{Streamlining Cross-Document Coreference Resolution:\\ Evaluation and Modeling}

\author{Arie Cattan\textsuperscript{1} \quad
    Alon Eirew\textsuperscript{1,2} \quad
    Gabriel Stanovsky\textsuperscript{3}\thanks{\; Work done while at the Allen Institute for AI and the University of Washington.} \quad
    Mandar Joshi\textsuperscript{4} \quad
    Ido Dagan\textsuperscript{1}\\ \\
\textsuperscript{1}Computer Science Department, Bar Ilan University \\ 
\textsuperscript{2}Intel Labs, Israel \quad
\textsuperscript{3}The Hebrew University of Jerusalem 
 \\
\textsuperscript{4}Allen School of Computer Science \& Engineering, University of Washington, Seattle, WA
 \\
  {\tt  arie.cattan@gmail.com} \quad {\tt  alon.eirew@intel.com} \\ 
  {\tt gabis@cse.huji.ac.il} \quad
     {\tt mandar90@cs.washington.edu} \\ {\tt  dagan@cs.biu.ac.il}  
  }

\date{}

\begin{document}
\maketitle

\begin{abstract}

Recent evaluation protocols for \emph{Cross-document} (CD) coreference resolution have often been inconsistent or lenient, leading to incomparable results across works and overestimation of performance. 
To facilitate proper future research on this task, our primary contribution is proposing a pragmatic evaluation methodology which assumes access to only \emph{raw} text -- rather than assuming gold mentions, disregards singleton prediction, and addresses typical targeted settings in CD coreference resolution. Aiming to set baseline results for future research that would follow our evaluation methodology, we build the first end-to-end model for this task.\footnote{\url{https://github.com/ariecattan/coref}} Our model adapts and extends recent neural models for within-document coreference resolution to address the CD coreference setting, which outperforms state-of-the-art results by a significant margin. 

\end{abstract}

\section{Introduction}
\begin{table*}
    \centering
    \resizebox{1\textwidth}{!}{
    \begin{tabular}{p{8cm}||p{8cm}}\toprule
    \multicolumn{1}{c}{\textbf{Subtopic 1}} &
    \multicolumn{1}{c}{\textbf{Subtopic 2}} \\
    \toprule
    \textbf{Doc 1} & \textbf{Doc 1} \\
    
    \emph{News that \textbf{\textcolor{blue}{Barack Obama}} may \textbf{\underline{\textcolor{cadmiumorange}{name}}} \textbf{\textcolor{red}{Dr. Sanjay Gupta}} of Emory University and CNN as his Surgeon General has caused a spasm of celebrity reporting.} 
    & 
    \emph{\textbf{\textcolor{blue}{President Obama}} will \textbf{\underline{\textcolor{darkspringgreen}{name}}} \textbf{\textcolor{chamoisee}{Dr. Regina Benjamin}} as U.S. Surgeon General in a Rose Garden announcement late this morning.} \\ \\
    
    \textbf{Doc 2} & \textbf{Doc 2} \\
    \emph{CNN's management confirmed yesterday that \textbf{\textcolor{red}{Dr. Gupta}} had been  \textbf{\underline{\textcolor{cadmiumorange}{approached}}} by the Obama team. } 
    &
    \emph{\textbf{\textcolor{blue}{Obama}} \textbf{\underline{\textcolor{darkspringgreen}{nominates}}}  new surgeon general: \textbf{\textcolor{chamoisee}{MacArthur ``genius grant" fellow Regina Benjamin. }}} 
    \\
    
    \bottomrule
    \end{tabular}}
    \caption{Example of sentences of topic 34 in ECB+: The underlined words represent events, same color represents a coreference cluster.
    Different documents describe the same event using different words (e.g name, approached). The two subtopics present a challenging case of ambiguity 
    between the two different \emph{nominations}. 
    }
    \label{tab:subtopic}
\end{table*}

The literature on coreference resolution has traditionally divided the task into two different settings, addressing the task at either the \emph{Within-document} (WD) or \emph{Cross-document} (CD) level. Each setting has presented different challenges, model design choices, and historically different evaluation practices.

In CD coreference resolution, the instances consist of multiple documents, each authored independently, without any inherent linear ordering between them.
As a result, coreferring expressions across documents are often lexically-divergent, while lexically-similar expressions may refer to different concepts. Table~\ref{tab:subtopic} shows example documents discussing similar, yet distinct, events (two different nominations of a US Surgeon General) with overlapping participants (``President Barack Obama'') and event triggers (``name''). 
Leveraging accurate CD coreference models seems particularly appealing for applications that merge information across texts, which have been gaining growing attention recently, such as multi-document summarization~\cite{falke-etal-2017-concept} and multi-hop question answering \cite{dhingra-etal-2018-neural, wang-etal-2019-multi-hop}.

In this paper, we observe that research on CD coreference has been lagging behind the impressive strides made in WD coreference~\cite{lee-etal-2017-end, lee2018higher, joshi-etal-2019-bert, doi:10.1162/tacl_a_00300, wu-etal-2020-corefqa}. 
As the time seems ripe to promote advances in CD coreference modeling as well, we present two steps to facilitate and trigger such systematic research, with respect to proper evaluation methodologies and current modeling approaches.

With respect to evaluation, we find that previous works have often used incomparable or lenient evaluation protocols, such as assuming event and entity mentions are given as part of the input, peeking into the fine-grained subtopic annotations, or rewarding coreference models for just identifying singleton clusters~(Section~\ref{sec:background}). As we will show in Section~\ref{sec:results}, these evaluation protocols have resulted in artificially inflated performance measures.

To address these shortcomings, our primary contribution consists of formalizing a realistic evaluation methodology for CD coreference. Namely, we use only raw input texts without assuming access to human-labeled annotations such as entity and event mentions, and also disregard singletons during evaluation. In addition, we examine model performance in both focused topic clusters, known a-priory to discuss overlapping information, as well as on larger sets of documents which contain both related and unrelated documents~(Section~\ref{sec:methodology}).


With respect to modeling, in Section~\ref{sec:model}, we describe a first end-to-end CD coreference model which builds upon the state-of-the-art in WD coreference and recent advances in transformer-based encoders. To achieve this, we address the inherently non-linear nature of the CD setting by combining this model with an agglomerative clustering approach, which was shown useful in other CD models. We first show that this combination sets a new state of the art for the task of CD coreference, in comparison to prior evaluations~(Section~\ref{sec:results}). 
We then evaluate this model following our realistic and more challenging evaluation methodology, setting a proper baseline for future research.

Taken together, our work brings the task of cross-document coreference resolution up to modern NLP standards, providing standardized evaluation benchmarks and a modern model which sets a new state-of-the-art result for the task. 
We hope that future work will use our framework to develop, and particularly to evaluate, models which make further advances on this challenging and important task.


\section{Background: Datasets, Evaluation, and Models}
\label{sec:background}

We first describe the problem of within-document~(WD) coreference as a reference point, in terms of established benchmark datasets, evaluation protocols, and state-of-the-art models~(Section~\ref{subsec:wd}).
In Section~\ref{subsec:cd} we similarly describe the current status of cross-document~(CD) coreference, which, as opposed to the WD setting, suffers from non-standard evaluation protocols and somewhat outdated models.

\subsection{Within-Document Coreference} 
\label{subsec:wd}
\paragraph{Benchmark dataset and evaluation}
OntoNotes~\cite{pradhan-etal-2012-conll} is the standard dataset for training and testing models in the WD setting. Each document in this corpus is exhaustively annotated with both event and entity coreference clusters, while omitting \emph{singletons} --- entities or events which are only mentioned once and are not co-referred to in the document. 
The OntoNotes coreference task formulation is designed to evaluate a model's ability to correctly identify mentions of coreferring entities and events, as well as the coreference links between them, given only the \emph{raw} document text.

\paragraph{State-of-the-art models}
Models for WD coreference resolution have closely followed and adopted to recent trends in NLP, converging on end-to-end deep learning architectures which do not require intermediate structure (e.g., syntactic trees) or task-specific processing.
\citet{lee-etal-2017-end} presented the first prominent work to introduce recurrent neural network for the task, significantly outperforming previous works, without requiring any additional resources beside task supervision. 
Successive follow-up works kept improving performance through the incorporation of widely popular pretrained architectures~\cite{lee2018higher}, culminating recently in the introduction of the now ubiquitous BERT model~\cite{devlin-etal-2019-bert} to WD coreference, thus achieving the current state of the art for the task~\cite{joshi-etal-2019-bert,kantor-globerson-2019-coreference,wu-etal-2020-corefqa}.

\subsection{Cross-Document Coreference} 
\label{subsec:cd}
\paragraph{Benchmark dataset}

The largest dataset that includes both WD and CD coreference annotation is ECB+~\cite{cybulska-vossen-2014-using}, which in recent years served as the main benchmark for CD coreference.
Each instance in ECB+ is a set of documents, dubbed a \emph{topic}, which consists of news articles in English (see Appendix~\ref{app:dataset} for more details).
To ensure that every instance poses challenging lexical ambiguity, each topic is a union of two sets of documents discussing two \emph{different} events (each called a \emph{subtopic}), yet which are likely to use a \emph{similar} vocabulary.
For example, Table~\ref{tab:subtopic} shows two fragments from the ``Obama's announcement of Surgeon General'' topic, one pertaining to the nomination of Dr. Sanjay Gapta, while the other discusses the nomination of Dr. Regina Benjamin, thus presenting a challenging disambiguation task.
While relatively small, this corpus represents a realistic use case for coreference detection across a restricted set of topically-related documents (e.g search results or multi-document summarization).

\paragraph{Evaluation}
The evaluation of models for CD coreference has commonly been more lenient, and less standardized than WD coreference, leading to incomparable results.
This stems from three major reasons.
First, while WD coreference requires models to identify entities, events, and their respective coreference links, evaluations of CD coreference mostly assumed that gold event and entity mentions are given as part of the input\footnote{\citet{yang-etal-2015-hierarchical, choubey-huang-2017-event} deviate from this setup and report results on raw text, yet consider only the intersection between gold and predicted mentions, not penalizing models for false positive mention identification.} \cite{cybulska-vossen-2015-translating, kenyon-dean-etal-2018-resolving, barhom-etal-2019-revisiting}.
Second, singleton clusters, as discussed in Section~\ref{subsec:wd-standard}, are \emph{not} excluded and constitute an integral part of the evaluation.\footnote{\citet{lee-etal-2012-joint} evaluated without singletons on EECB, but subsequent works did not follow their methodology.}
Finally, CD models have been inconsistent with their usage of topic and subtopic information. Some works have evaluated performance on gold \emph{subtopics}
\cite{yang-etal-2015-hierarchical, choubey-huang-2017-event}, thus obviating the aforementioned designed lexical ambiguity at the topic level~\cite{upadhyay-etal-2016-revisiting}.
Recent models~\cite{barhom-etal-2019-revisiting, Meged2020ParaphrasingVC} apply as preprocessing, a document clustering on all the corpus. However, due to the high lexical similarity between documents within the same subtopic, this yields an almost perfect \emph{subtopic} clustering. Considering the aforementioned nature of ECB+, such clustering can be regarded as evaluation at the subtopic level, and should be avoided.
Furthermore, evaluating only on individual subtopics disregards the fact that a coreference cluster may involve two subtopics, as we can see in Table~\ref{tab:subtopic} where ``Barack Obama" appears in two different subtopics.

\paragraph{State-of-the-art models}

Unlike WD coreference, models for CD coreference seem to be behind the curve of recent NLP advances, mostly using hand-crafted features~\cite{cybulska-vossen-2015-translating, yang-etal-2015-hierarchical, choubey-huang-2017-event, kenyon-dean-etal-2018-resolving}.
Recently,~\citet{barhom-etal-2019-revisiting} proposed to jointly address event and entity coreference as a single task. For that purpose, they simulate the clustering process during training, and recalculate new mention representations and pairwise scores after each cluster merging step. Based on this model, \citet{Meged2020ParaphrasingVC} improved results on event coreference by leveraging a paraphrase resource~(Chirps;~\citealp{shwartz-etal-2017-acquiring}) as distant supervision. While computationally complexs and relying on many external resources, these methods outperform single lexical match baselines by relatively small margins.

\section{Proposed Evaluation Methodology}
\label{sec:methodology}

In this section, we propose a standard evaluation methodology for CD coreference which addresses its main limitations, described in the previous section. 
First, in Section~\ref{subsec:wd-standard} we propose to be consistent with the WD task formulation \cite{pradhan-etal-2012-conll} by:
(1) assuming raw textual input without gold mention annotations; 
and (2) omitting singletons from the evaluation.
In addition, our evaluation protocol proposes a standard break down performance at topic and corpus level~(Section~\ref{subsec:topic}), thus standardizing the previously non-comparable usage of topic and subtopic information.

\subsection{Adapted Single-Document Standard}
\label{subsec:wd-standard}

\paragraph{Raw documents as input}
We argue that CD coreference models should be mainly evaluated on raw text input, without assuming access to gold entity and event annotations. 
That is, models should perform coreference clustering of predicted rather than gold mentions. 
This setup is closer to the most recent NLP task's formulation, and while being significantly more challenging, it simulates real-world use-cases. 
Evaluating on gold mentions can still be valuable for error analysis, i.e., analyzing the degree to which a model erred because of incorrect mention identification vs. because of incorrect coreference linking.

\paragraph{Omitting singletons from the evaluation}

Similar to common practice in WD coreference, we propose to omit singleton clusters from the CD evaluation process.
In fact, a model's ability to identify that singletons do \emph{not} belong to any coreference cluster is already captured in the coreference evaluation metrics. However, as we further analyze in Appendix~\ref{app:singleton}, including singletons during the evaluation distorts the measurement of the mention-based metrics B\textsuperscript{3}~\cite{bagga-baldwin-1998-entity}, CEAFe~\cite{luo-2005-coreference} and LEA~\cite{moosavi-strube-2016-coreference} by rewarding (or penalizing) identification of singleton span boundaries. As singletons constitute a major part of the text, this will bias the results towards models that perform well on detecting all the mentions, rather than the coreference links. Such evaluation is counterproductive with the downstream goal of coreference resolution — providing cross-mention links which enable reasoning over distributed information, while singletons are typically not relevant for this downstream purpose.

When evaluating only on gold mentions, including singletons further harms the validity of the current CD evaluation protocol and artificially inflates the results. Evidently, a dummy baseline which predicts no coreference links and puts each input gold mention in a singleton cluster achieves non-negligible performance \cite{luo-2005-coreference} (see Tables~\ref{tab:subtopic_results_event} and~\ref{tab:subtopic_results_entity}).

\subsection{Topic and Corpus Level Evaluation}
\label{subsec:topic}

As mentioned in Section~\ref{sec:background}, CD coreference models have previously made inconsistent usage of topic and subtopic information.
We address this by breaking down CD model evaluation to two settings:

\paragraph{Corpus level performance:}  

An input instance in this setting consists of a single set of documents, omitting information about the different topics and subtopics (e.g exact number of topics)~\cite{cybulska-vossen-2015-translating, upadhyay-etal-2016-revisiting, kenyon-dean-etal-2018-resolving, barhom-etal-2019-revisiting}. This evaluation does not make any assumption about the dataset and is also suitable for corpora in which the documents are not categorized into topics.
    
\paragraph{Topic level performance:} Here, each gold \emph{topic} is evaluated separately~\cite{bejan-harabagiu-2010-unsupervised}. In this setting, an input instance consists of a set of documents pertaining to a single topic, including, in the case of ECB+, the two subtopics which present a challenging lexical ambiguity.  
While this setup makes the coreference task simpler than the corpus level evaluation~\cite{upadhyay-etal-2016-revisiting}, it simulates a realistic scenario where documents are initially collected at the topic level.
For example, in multi-document summarization, where the goal is to generate a short summary of a topic including several documents, applying coreference resolution on the input documents has been shown to be useful for merging similar concepts~\cite{falke-etal-2017-concept} and generating coherent summaries~\cite{christensen-etal-2013-towards}.

\section{Model}
\label{sec:model}

\begin{figure*}[!ht]
\centering
\includegraphics[width=\textwidth]{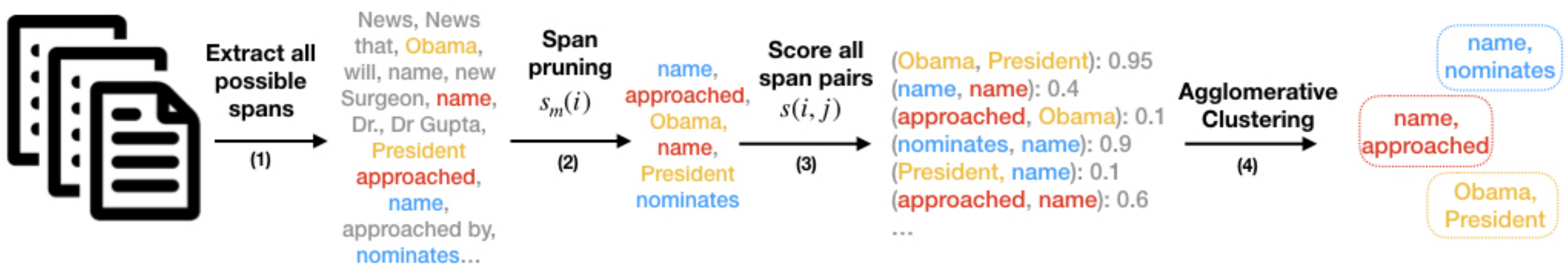}
\caption{Overall model flow, with examples from Table~\ref{tab:subtopic}. 
(1) extract and score all possible spans (2) keep top spans according to $s_m(i)$ (3) score all pairs $s(i, j)$ and (4) cluster spans using agglomerative clustering.
\label{fig:diagram}
}\end{figure*}

To establish baseline results on ECB+ with predicted mentions, we built an end-to-end CD coreference model, inspired by the successful \emph{e2e-coref} single-document coreference model~\cite{lee-etal-2017-end}, which jointly learns mention detection and coreference link prediction, as elaborated in Section~\ref{subsec:e2e}. 
We modify its clustering method and training objective to port it to the cross-document setting (Section~\ref{subsec:newmodel}).

\subsection{Overview of e2e-coref}
\label{subsec:e2e}

For each span $i$, the model learns a distribution $P(\cdot)$ over its possible antecedent spans $Y(i)$:
\begin{align*}
    P(y_i) = \frac{e^{s(i, y_i)}}{\sum_{y^{\prime} \in Y(i)} e^{s(i, y^{\prime})}}
\end{align*}
Considering all spans in a document as potential mentions, the scoring function $s(i, j)$ between span $i$ and $j$, where $j$ appears before $i$, has three components: the two mention scores $s_m(\cdot)$ of spans $i$ and $j$, and a pairwise score $s_a(i, j)$ for span $j$ being antecedent of span $i$. 

After encoding all the tokens in a document, each possible span up to a length $n$ is represented with the concatenation of four vectors: the output representations of the span boundary (first and last) tokens, an attention-weighted sum of token representations in the span $\hat{x}_i$, and a feature vector $\phi(i)$ denoting the span length.
These span representations are first fed into a mention scorer $s_m(\cdot)$ to filter the $\lambda T$ (where $T$ is the number of tokens in the document) spans with the highest scores.
Then, the model learns for each of these spans to optimize the marginal log-likelihood of its correct antecedents. The full description of the different components is described below:
\begin{align*} 
\label{span_repr}
s(i, j) &= s_m(i) + s_m(j) + s_a(i, j) \\
s_{m}(i) &= \mathnormal{ w_m \cdot \text{FFNN}_m(g_i)} \\
s_{a}(i, j) &= w_a \cdot \text{FFNN}_a([g_i, g_j, g_i \circ g_j]) \\
g_{i} &= [x_\text{FIRST(i)}, x_\text{LAST(i)}, \hat{x}_i, \phi(i) ]
\end{align*}

\subsection{End-to-end Cross-Document Coreference}
\label{subsec:newmodel}

The major obstacle in applying the \emph{e2e-coref} model directly in the CD setting is its reliance on \emph{textual ordering} --
it forms coreference chains by linking each mention to an antecedent span appearing before it in the document. This clustering method cannot be used in the multiple-document setting since there is no inherent ordering between the documents. 

\paragraph{Clustering Spans} To overcome this challenge, we combine the model architecture from \emph{e2e-coref} with an agglomerative clustering-based approach, as common in CD coreference resolution~\cite{yang-etal-2015-hierarchical, choubey-huang-2017-event, kenyon-dean-etal-2018-resolving, barhom-etal-2019-revisiting}. 
The overall architecture of our model is shown in~Figure~\ref{fig:diagram}.
The agglomerative clustering step merges the most similar cluster pairs until their pairwise similarity score falls below a tuned threshold $\tau$.
Following the average-link method, the cluster pair score is defined as the average of span pair similarity scores $s(i, j)$ (from the \emph{e2e-coref} architecture) over all span pairs $(i, j)$ across the two candidate clusters to be merged.

\paragraph{Training}
We train the model by optimizing a binary cross-entropy loss over pairs of mentions.  
Specifically, given a set of documents, the first step consists of encoding each document separately using RoBERTa\textsubscript{\emph{LARGE}}~\cite{liu2019roberta}. 
Long documents are split into non overlapping segments of up to 512 word pieces tokens and are encoded independently \cite{joshi-etal-2019-bert}. 
For computation efficiency, we prune spans greedily, keeping only the $\lambda T$ (where $T$ is the total number of tokens in all the documents) highest scoring spans according to the mention scorer $s_m(\cdot)$. 
Unlike the \emph{e2e-coref}, we pre-train the mention scorer $s_m(\cdot)$ on the ECB+ gold mention spans.
Next, instead of comparing a mention only to its previous spans in the text, our pairwise scorer $s(i, j)$ compares a mention to all other spans across all the documents.\footnote{In practice, since the documents in ECB+ are rather short (Appendix~\ref{app:dataset}), these pairs are mostly composed of spans from different documents.} The positive instances for training consist of all the pairs of mentions that belong to the same coreference cluster, while the negative examples are sampled (20x the number of positive pairs) from all other pairs.
Due to memory constraint, we freeze output representations from RoBERTa instead of fine-tuning all parameters. The mention scorer $s_m(\cdot)$ and the pairwise scorer $s_a(i, j)$ are jointly learned to optimize the binary cross-entropy loss as follows:

\begin{align*}
    L = -\frac{1}{|N|} \sum_{x, z \in N} {y \cdot log(s(x, z))}
\end{align*}
where N corresponds to the set of mention-pairs, and $y \in \{0, 1\}$ to a pair label. Full implementation details are described in Appendix~\ref{app:model}. When training and evaluating the model using gold mentions, we ignore the span mention scores, $s_m(\cdot)$, and the gold mention representations are directly fed into the pairwise scorer $s_a(i, j)$.

\paragraph{Inference}
At inference time, we score all spans; prune spans with lowest scores; score the pairs; and finally form the coreference clusters using an agglomerative clustering over these pairwise scores. 

\paragraph{Topic Level Processing}
To limit the search space, we apply the above algorithm separately for each \emph{topic} (cluster of documents).
During training, we use the gold topic segmentation of the training data. At inference time, we construct the set of topics differently for topic and corpus level evaluation (Section ~\ref{subsec:topic}). We use the gold topics when evaluating at the topic level, as each topic is evaluated independently. However, for corpus level, since this evaluation protocol assumes the number of topics is unknown, we predict the \emph{topic} clusters using another agglomerative clustering over the document representations until the document similarity drops below a threshold.
Specifically, the documents are represented using TF-IDF scores of unigrams, bigrams, and trigrams, and they are merged according to their cosine similarity.

\begin{table*}[!h]
    \centering
    \resizebox{\textwidth}{!}{
    \begin{tabular}{@{}lcccccccccccccccclc@{}}
    \toprule
    &\phantom{abcd}& \multicolumn{3}{c}{MUC} && \multicolumn{3}{c}{$B^3$} & & \multicolumn{3}{c}{$CEAFe$} && \multicolumn{3}{c}{LEA} && CoNLL\\
    \cmidrule{3-5} \cmidrule{7-9} \cmidrule{11-13} \cmidrule{15-17} \cmidrule{19-19}
    && R & P & $F_1$ && R & P & $F_1$ && R &P & $F_1$ && R &P & $F_1$ && $F_1$  \\ 
   \midrule
        Singleton$^+$ && 0 & 0 & 0 && 45.2 & 100 & 62.3 &&  86.7 & 39.2 & 54.0 && 35.0 & 35.0 & 35.0 && 38.8 \\
        \vspace{1.7mm}
        Singleton$^-$ && 0 & 0 & 0 && 0 & 0 & 0 && 0 & 0 & 0 &&0&0&0 && 0\\
        Same Head-Lemma$^+$ &&  76.5 & 80.0 & 78.2 && 71.8 & 85.0 & 77.8 && 75.5 & 71.8 & 73.6 && 57.0 & 72.0 & 63.7 && 76.5 \\
        \vspace{1.7mm}
        Same Head-Lemma$^-$ &&  76.5 & 80.0 & 78.2 && 54.4 & 72.6 & 62.2 && 68.0 & 42.5 & 52.3 && 50.6 & 69.6 & 58.6 && 64.2\\
        
        \citet{barhom-etal-2019-revisiting}$^+$ && 78.1 & 84.0 & 80.9 && 76.8 & 86.1 & 81.2 && 79.6 & 73.3 & 76.3 && 64.6 & 72.3 & 68.3 && 79.5  \\
        \vspace{1.7mm}
        \citet{barhom-etal-2019-revisiting}$^-$ && 78.1 & 84.0 & 80.9 &&  61.2 & 73.5 & 66.8 && 63.2 & 48.9 & 55.2 && 58.4 & 71.2 & 64.2 && 67.6  \\
        
        \citet{Meged2020ParaphrasingVC}$^+$ && 78.8 & 84.7 & 81.6 && 75.9 & 85.9 & 80.6 && 81.1 & 74.8 & 77.8 && 64.7 & 73.4 & 68.8 && 80.0  \\
        \citet{Meged2020ParaphrasingVC}$^-$ && 78.8 & 84.7 & 81.6 && 60.4 & 73.8 & 66.4 && 65.5 & 49.5 & 56.4 && 57.2 & 71.2 & 63.4 && 68.1 \\
        
        \midrule
        Our model$^+$ &&  85.1 & 81.9 & 83.5 && 82.1 & 82.7 & 82.4 && 75.2 & 78.9 & 77.0 && 68.8 & 72.0 & 70.4 && \textbf{81.0}   \\
        Our model$^-$ && 85.1 & 81.9 & 83.5 && 70.8 & 70.2 & 70.5 && 68.2 & 52.3 & 59.2 && 68.2 & 67.6 & 67.9 && \textbf{71.1}   \\
    \bottomrule
    \end{tabular}}
    \caption{Event coreference on ECB+ test, on predicted subtopics and gold mentions, with($^+$)/without($^-$) singletons}
    \label{tab:subtopic_results_event}
\end{table*}

\begin{table*}[!h]
    \centering
    \resizebox{\textwidth}{!}{
    \begin{tabular}{@{}lcccccccccccccccclc@{}}
    \toprule
    &\phantom{abcd}& \multicolumn{3}{c}{MUC} && \multicolumn{3}{c}{$B^3$} & & \multicolumn{3}{c}{$CEAFe$} && \multicolumn{3}{c}{LEA} && CoNLL\\
    \cmidrule{3-5} \cmidrule{7-9} \cmidrule{11-13} \cmidrule{15-17} \cmidrule{19-19}
    && R & P & $F_1$ && R & P & $F_1$ && R &P & $F_1$ && R &P & $F_1$ && $F_1$  \\ 
    \midrule
    Singleton$^+$ && 0 & 0 & 0 && 29.6 & 100 & 45.7 &&  80.3 & 23.8 & 36.7 && 20.1 & 20.1 & 20.1 && 27.5 \\
        \vspace{1.7mm}
        Singleton$^-$ && 0 & 0 & 0 && 0 & 0 & 0 && 0 & 0 & 0 &&0&0&0 && 0\\
        Same Head-Lemma$^+$ &&   71.3 & 83.0 & 76.7 && 53.4 & 84.9 & 65.6 && 70.1 & 52.5 & 60.0 && 40.6 & 69.1 & 51.1 && 67.4 \\
        \vspace{1.7mm}
        
        Same Head-Lemma$^-$ &&  71.3 & 83.0 & 76.7 &&  39.4 & 77.2 & 52.2 && 60.1 & 34.7 & 44.0 && 35.2 & 73.6 & 47.6 && 57.6 \\
        \citet{barhom-etal-2019-revisiting}$^+$ &&  81.0 & 80.8 & 80.9 && 66.8 & 75.5 & 70.9 && 62.5 & 62.8 & 62.7 && 53.5 & 63.8 & 58.2 && 71.5 \\
        \citet{barhom-etal-2019-revisiting}$^-$ && 81.0 & 80.8 & 80.9 &&  57.3 & 67.3 & 61.9 && 60.4 & 42.1 & 49.6 && 54.1 & 63.5 & 58.5 && 64.1 \\
        \midrule
        Our model$^+$ &&  85.7 & 81.7 & 83.6 && 70.7 & 74.8 & 72.7 && 59.3 & 67.4 & 63.1 && 56.8 & 65.8 & 61.0 && \textbf{73.1}  \\
        Our model$^-$ &&  85.7 & 81.7 & 83.6 && 62.4 & 67.6 & 64.9 && 62.3 & 46.6 & 53.3 && 59.3 & 65.0 & 62.0 && \textbf{67.3}  \\
    \bottomrule
    \end{tabular}}
    \caption{Entity coreference on ECB+ test, on predicted subtopics and gold mentions, with($^+$)/without($^-$) singletons}
    \label{tab:subtopic_results_entity}
\end{table*}

\section{Empirical Assessments and Results}
\label{sec:results}

\subsection{Empirical Assessments }
\label{inflation}

To assess the effectiveness of our model, in comparison to prior models, we first evaluate it under the current common evaluation setting. Specifically, we use gold mentions as input and cluster the documents into subtopics, as done in~\cite{barhom-etal-2019-revisiting}. We compare our model with the same head-lemma baseline\footnote{This baseline merges mentions sharing the same syntactic-head-lemma into a coreference cluster.} and two recent neural state-of-the-art models~\cite{barhom-etal-2019-revisiting,Meged2020ParaphrasingVC}. We did not compare to~\citet{yang-etal-2015-hierarchical} and~\citet{choubey-huang-2017-event} because they use a different ECB+ setup with known annotation errors, as already criticized in~\cite{barhom-etal-2019-revisiting}.
Also, in order to estimate the artificial inflation caused by the inclusion of singletons, we add the singleton baseline and re-evaluate all baselines also while omitting singletons, as well as adding the recent LEA metric~\cite{moosavi-strube-2016-coreference}. The results are reported in Table~\ref{tab:subtopic_results_event} for events and in Table~\ref{tab:subtopic_results_entity} for entities.

The results using the MUC evaluation measure remain identical after removing singletons, because MUC is a link-based metric which ignores singletons. Although LEA is also a \emph{link-based} metric, it handles singletons by self-links, and the results differ accordingly when omitting singletons.
On the other hand, the results using the \emph{mention-based} metrics B\textsuperscript{3} and CEAFe are significantly higher when including singletons (e.g +11.9 B\textsuperscript{3} F1 and +17.8 CEAFe F1 for our model on event coreference). Indeed, as explained in Section~\ref{subsec:wd-standard}, these metrics unjustifiably reward models for correctly predicting \emph{gold} mention spans. This inflation also stems from the \emph{mention identification effect}~\cite{moosavi-strube-2016-coreference}, which is amplified when using gold mentions.


Overall, our model offers an improvement of 3 F1 points in event and 3.2 F1 points in entity coreference when ignoring singletons over the current state-of-the-art models~\cite{barhom-etal-2019-revisiting, Meged2020ParaphrasingVC} on ECB+, while surpassing the strong lemma baseline by 6.9 and 9.7 points respectively. 
Beyond improving the results, our model is simpler as it does not rely on external resources such as SRL, WD coreference resolver, or a paraphrase resource. Also, it is more efficient in both training and inference since it computes pairwise scores using a simple MLP only once.

\subsection{Results}

\begin{table*}
    \centering
    \resizebox{\textwidth}{!}{
    \begin{tabular}{lllccclccclccclccclc}
    \toprule
    \phantom{defgrnbkjb} & \phantom{defgrnbkjb} && \multicolumn{3}{c}{MUC} && \multicolumn{3}{c}{B\textsuperscript{3}} && \multicolumn{3}{c}{$CEAFe$} &&
    \multicolumn{3}{c}{LEA} &&
    CoNLL\\
    \cmidrule{4-6} \cmidrule{8-10} \cmidrule{12-14} \cmidrule{16-18} \cmidrule{20-20}
    &&& R & P & $F_1$ && R & P & $F_1$ && R &P & $F_1$ && R &P & $F_1$ && $F_1$  \\
    \midrule
    \multirow{2}{*}{Event} & Topic && 64.8 & 62.0 & 63.4 && 52.5 & 36.8 & 43.2 && 40.2 & 36.2 & 38.1 && 49.6 & 33.4 & 40.0 && 48.2 \\
    & Corpus && 64.3 & 61.0 & 62.6 && 51.5 & 34.4 & 41.2 && 38.3 & 35.5 & 36.8 && 48.5 & 31.0 & 37.8 && 46.9 \\
    \midrule
    \multirow{2}{*}{Entity} & Topic && 41.7 & 52.3 & 46.4 && 24.8 & 37.1 & 29.7 && 27.4 & 26.8 & 27.1 && 22.3 & 34.4 & 27.1 && 34.4 \\
    & Corpus && 41.7 & 52.3 & 46.4 && 24.6 & 37.0 & 29.6 && 27.2 & 26.1 & 26.7 && 22.3 & 34.3 & 27.0 && 34.2 \\
    \midrule
    \multirow{2}{*}{ALL} & Topic && 49.3 & 56.7 & 52.8 && 31.7 & 41.4 & 35.9 && 39.2 & 32.4 & 35.5 && 28.7 & 37.5 & 32.5 && 41.4 \\
    & Corpus && 49.3 & 55.9 & 52.4 && 31.5 & 39.2 & 34.9 && 37.1 & 32.8 & 34.8 && 28.6 & 35.5 & 
    31.7 && 40.7 \\
    \bottomrule
    \end{tabular}}
    \caption{Results of our model on the ECB+ test set, predicted mentions, w/o singletons, topic and corpus level.}
    \label{tab:predicted_mentions}
\end{table*}

\begin{table*}
    \centering
    \resizebox{\textwidth}{!}{
    \begin{tabular}{lllccclccclccclccclc}
    \toprule
    \phantom{defgrnbkjb} & \phantom{defgrnbkjb} && \multicolumn{3}{c}{MUC} && \multicolumn{3}{c}{B\textsuperscript{3}} && \multicolumn{3}{c}{$CEAFe$} &&
    \multicolumn{3}{c}{LEA} &&
    CoNLL\\
    \cmidrule{4-6} \cmidrule{8-10} \cmidrule{12-14} \cmidrule{16-18} \cmidrule{20-20}
    &&& R & P & $F_1$ && R & P & $F_1$ && R &P & $F_1$ && R &P & $F_1$ && $F_1$  \\
    \midrule
    \multirow{2}{*}{Event} & Topic && 80.1 & 76.3 & 78.1 && 63.4 & 54.1 & 58.4 && 56.3 & 44.2 & 49.5 && 59.7 & 49.6 & 54.2 && 62.0 \\
    & Corpus && 79.9 & 74.8 & 77.2 && 62.2 & 48.9 & 54.8 && 53.3 & 42.3 & 47.2 && 58.4 & 44.4 & 50.5 && 59.7 \\
    \midrule
    \multirow{2}{*}{Entity} & Topic && 85.8 & 79.3 & 82.4 && 64.3 & 60.0 & 62.1 && 58.6 & 45.9 & 51.5 && 60.9 & 56.8 & 58.8 && 65.3 \\
    & Corpus && 85.7 & 79.3 & 82.4 && 63.7 & 60.0 & 61.8 && 58.1 & 45.0 & 50.7 && 60.3 & 56.8 & 58.5 && 65.0  \\
    \midrule
    \multirow{2}{*}{ALL} & Topic && 83.8 & 78.4 & 81.0 && 66.0 & 55.5 & 60.3 && 56.2 & 43.3 & 48.9 && 62.6 & 51.8 & 56.7 && 63.4 \\
    & Corpus && 83.8 & 77.4 & 80.5 && 65.8 & 51.9 & 58.1 && 52.9 & 43.5 & 47.8 && 62.4 & 48.4 & 54.5 && 62.1 \\
    \bottomrule
    \end{tabular}}
    \caption{Results of our model on the ECB+ test set, gold mentions, w/o singletons, topic and corpus level.}
    \label{tab:gold_mentions}
\end{table*}

Here, we evaluate our model according to our proposed evaluation methodology~(Section~\ref{sec:methodology}), in order to set the state-of-the-art baseline performance for future work on ECB+.
The primary results are presented in Table~\ref{tab:predicted_mentions}, evaluated on predicted mentions. Additionally, Table~\ref{tab:gold_mentions} presents performance over gold mentions, allowing to analyze the impact of mention prediction. Per our methodology, in both tables results are presented for the topic and corpus levels, while ignoring singletons in the evaluation.

Since our model architecture is not tailored for a specific mention type, we use the same model separately for both event and entity coreference. In addition, inspired by \citet{lee-etal-2012-joint} and by the single-document standard, we train our model to perform event and entity coreference together as a single unified task, that we term ``ALL".\footnote{This approach is different from the \textsc{JOINT} model of \citet{barhom-etal-2019-revisiting}, which does distinguish between event and entity mentions at test time.} This represents a useful scenario when we are interested in finding all the coreference links in a set of documents. 
Addressing CD coreference with ALL is challenging because \emph{(i)} the search space is much larger, and \emph{(ii)} this involves subtle distinction for the model (e.g voters vs voted). 
For all experiments, we use a single GeForce GTX 1080 Ti 12GB GPU, where the training takes 2.5 hours for the most expensive setting (ALL on predicted mentions), while inference takes 11 minutes.

\subsection{Ablations}

In order to show the importance of each component of our model, we ablate several parts and compute F1 scores on the development set of the ECB+ event dataset. Results are presented in Table~\ref{tab:ablations} using predicted mentions at the topic level. 
\begin{table}[!h]
    \newcommand{\colindent}{\;}
    \centering
    \resizebox{0.48\textwidth}{!}{
    \begin{tabular}{lcc}
    \toprule
    \phantom{fwidsvhckzxjvchndfzxvvgdaczfc} & F1 & $\Delta$\\
    \midrule
    Our model  & 58.1 & \\
    \colindent $-$ pre-train of mention scorer & 54.9 & -3.2 \\
    \colindent $-$ dynamic pruning & 54.1 & -4.0 \\
    \colindent $-$ RoBERTa & 54.0 & -4.1\\
    \colindent $-$ negative sampling & 56.7 & -1.4 \\
    \bottomrule
    \end{tabular}}
    \caption{Ablation results (CoNLL F1) on our model on the development set of ECB+ event coreference on the topic level. Pre-training of the mention scorer, the dynamic pruning, RoBERTa, and negative sampling, all contribute significantly to the model performance. }
    \label{tab:ablations}
\end{table}

\paragraph{Pre-training of mention scorer}
Skipping the pre-training of the mention scorer results in a 3.2 F1 points degradation in performance. Indeed, the relatively small training data in the ECB+ dataset (see Appendix~\ref{app:dataset}) might be not sufficient for using only end-to-end optimization, and pre-training of the mention scorer helps generate good candidate spans from the first epoch. 

\paragraph{Dynamic pruning} To analyze the effect of the dynamic pruning, we froze the mention scorer during the pairwise training, and keep the same candidate spans along the training. The significant drop in performance (4.0 F1 points) reveals that the mention scorer inherently incorporates coreference signals. 

\paragraph{RoBERTa} 
Replacing RoBERTa\textsubscript{\emph{LARGE}} with vanilla BERT\textsubscript{\emph{LARGE}}~\cite{devlin-etal-2019-bert} decreases the results by 4.1 points. This is in line with the powerful of RoBERTa scores over BERT on various tasks. 

\paragraph{Negative Sampling}
Using all negative pairs for training leads to a performance drop of 1.4 and significantly increases the training time.

\subsection{Analysis}

\paragraph{Subtopic, topic, corpus} 
Already when evaluating on gold mentions, the performance is much lower at the \emph{topic} level (Table \ref{tab:gold_mentions}) than at the \emph{subtopic} level~(Tables~\ref{tab:subtopic_results_event} and~\ref{tab:subtopic_results_entity}). 
Considering the nature of ECB+ where each topic consists of two similar subtopics describing two different news events~(Section~\ref{sec:background}), evaluating at the subtopic level removes this designed ambiguity challenge and should therefore be avoided. This aspect also explains why the performance drop is more substantial in event coreference  (71.1 vs. 62.0) than in entity (67.3 vs. 65.3), since the subtopics are based on event similarity. 
However, we see a slight performance gap between the \emph{topic} and \emph{corpus} level evaluation. For example, the model trained on event coreference achieves 48.2 F1 on the topic level and 46.9 on the corpus level. This demonstrates that our topic clustering algorithm, which precedes the coreference resolution step, achieves a reasonable segmentation of the documents to topics---reducing the search space without a major drop in performance. This algorithm clustered the 10 topics of the test set into 8 predicted topics. Although some gold topics were mixed, the pairwise scorer did manage to give relatively low scores to negative mention pairs across different topics.

\paragraph{Predicted mentions}
Overall, the performance on predicted mentions (main evaluation) is relatively lower (Table \ref{tab:predicted_mentions}) than that on gold mentions (Table~\ref{tab:gold_mentions}).
This performance drop is in harmony with the single-document setting, where using gold mentions was shown to offer an improvement of 17.5 F1 \cite{lee-etal-2017-end}, which corresponds to 50\% gain on error reduction. 
Therefore, additionally to the needed progress in making coreference decisions, there is also a large room for improvement in mention detection.        

\paragraph{Qualitative analysis}

We sampled topics from the development set and manually analyzed the errors of the ``ALL" configuration.
The most commonly occurring errors were due to an over reliance on lexical similarity. 
For example, the event ``\emph{Maurice Cheeks was \textbf{fired}}" was wrongly predicted to be coreferent with a similar, but distinct event, ``\emph{the Sixers \textbf{fired} Jim O'Brien}". 
On the other hand, the model sometimes struggles to merge mentions which are lexically different (e.g ``\emph{Jim O'Brien \textbf{was shown the door}}",  ``\emph{Jim O'Brien has been \textbf{relieved}}", ``\emph{Philadelphia \textbf{fire} coach Jim O'Brien}").

The model also seems to struggle with temporal reasoning, in part, due to missing information. For example, news articles from different days have different relative reference to time, while the publication date of the articles is not always available. As a result, the model did not link ``\emph{Today}" in one document to ``\emph{Saturday}" in another document, while both referred to the same day.

\section{Conclusion}

In this paper, we proposed a realistic evaluation methodology for \emph{cross-document} coreference resolution addressing the main shortcomings of current evaluation protocols.
Our proposal follows well-established standards in \emph{Within-document} coreference resolution. Models are mainly evaluated on \emph{raw} text while singletons are omitted during the evaluation. In addition, we formalize the usage of topic/subtopic segmentation during the evaluation for addressing the specific ambiguity challenges in CD coreference resolution. 
We also established the first end-to-end baseline for CD coreference resolution, with a simple and efficient model that does not rely on external resources.
Our model outperforms state-of-the-art results by 3 F1 points with respect to current evaluation methodologies.
To the best of our knowledge, this is also the first publicly released model for cross-document coreference resolution, which is easily applicable for downstream use over raw text.
Finally, we showed that when evaluating with our strict evaluation methodology, particularly when addressing the ambiguity of the corpus and topic levels (vs. sub-topics), performance dramatically decreases, suggesting a large room for improvement in future research. 



\section*{Acknowledgments}

We thank Shany Barhom for fruitful discussion and sharing code, and Yehudit Meged for providing her coreference predictions. This work was supported in part by grants from Intel Labs, Facebook, the Israel Science Foundation grant 1951/17, the Israeli Ministry of Science and Technology and the German Research Foundation through the German-Israeli Project Cooperation (DIP, grant DA 1600/1-1).

\bibliographystyle{acl_natbib}
\bibliography{anthology, emnlp2020}

\begin{thebibliography}{33}
\expandafter\ifx\csname natexlab\endcsname\relax\def\natexlab#1{#1}\fi

\bibitem[{Bagga and Baldwin(1998)}]{bagga-baldwin-1998-entity}
Amit Bagga and Breck Baldwin. 1998.
\newblock \href {https://doi.org/10.3115/980845.980859} {Entity-based
  cross-document core f erencing using the vector space model}.
\newblock In \emph{36th Annual Meeting of the Association for Computational
  Linguistics and 17th International Conference on Computational Linguistics,
  Volume 1}, pages 79--85, Montreal, Quebec, Canada. Association for
  Computational Linguistics.

\bibitem[{Barhom et~al.(2019)Barhom, Shwartz, Eirew, Bugert, Reimers, and
  Dagan}]{barhom-etal-2019-revisiting}
Shany Barhom, Vered Shwartz, Alon Eirew, Michael Bugert, Nils Reimers, and Ido
  Dagan. 2019.
\newblock \href {https://doi.org/10.18653/v1/P19-1409} {Revisiting joint
  modeling of cross-document entity and event coreference resolution}.
\newblock In \emph{Proceedings of the 57th Annual Meeting of the Association
  for Computational Linguistics}, pages 4179--4189, Florence, Italy.
  Association for Computational Linguistics.

\bibitem[{Bejan and Harabagiu(2008)}]{bejan-harabagiu-2008-linguistic}
Cosmin Bejan and Sanda Harabagiu. 2008.
\newblock \href
  {http://www.lrec-conf.org/proceedings/lrec2008/pdf/734_paper.pdf} {A
  linguistic resource for discovering event structures and resolving event
  coreference}.
\newblock In \emph{LREC 2008}.

\bibitem[{Bejan and Harabagiu(2010)}]{bejan-harabagiu-2010-unsupervised}
Cosmin Bejan and Sanda Harabagiu. 2010.
\newblock \href {https://www.aclweb.org/anthology/P10-1143} {Unsupervised event
  coreference resolution with rich linguistic features}.
\newblock In \emph{Proceedings of the 48th Annual Meeting of the Association
  for Computational Linguistics}, pages 1412--1422, Uppsala, Sweden.
  Association for Computational Linguistics.

\bibitem[{Chen et~al.(2018)Chen, Fan, Lu, Yuille, and
  Rong}]{chen-etal-2018-preco}
Hong Chen, Zhenhua Fan, Hao Lu, Alan Yuille, and Shu Rong. 2018.
\newblock \href {https://doi.org/10.18653/v1/D18-1016} {{P}re{C}o: A
  large-scale dataset in preschool vocabulary for coreference resolution}.
\newblock In \emph{Proceedings of the 2018 Conference on Empirical Methods in
  Natural Language Processing}, pages 172--181, Brussels, Belgium. Association
  for Computational Linguistics.

\bibitem[{Choubey and Huang(2017)}]{choubey-huang-2017-event}
Prafulla~Kumar Choubey and Ruihong Huang. 2017.
\newblock \href {https://doi.org/10.18653/v1/D17-1226} {Event coreference
  resolution by iteratively unfolding inter-dependencies among events}.
\newblock In \emph{Proceedings of the 2017 Conference on Empirical Methods in
  Natural Language Processing}, pages 2124--2133, Copenhagen, Denmark.
  Association for Computational Linguistics.

\bibitem[{Christensen et~al.(2013)Christensen, {Mausam}, Soderland, and
  Etzioni}]{christensen-etal-2013-towards}
Janara Christensen, {Mausam}, Stephen Soderland, and Oren Etzioni. 2013.
\newblock \href {https://www.aclweb.org/anthology/N13-1136} {Towards coherent
  multi-document summarization}.
\newblock In \emph{Proceedings of the 2013 Conference of the North {A}merican
  Chapter of the Association for Computational Linguistics: Human Language
  Technologies}, pages 1163--1173, Atlanta, Georgia. Association for
  Computational Linguistics.

\bibitem[{Cybulska and Vossen(2014)}]{cybulska-vossen-2014-using}
Agata Cybulska and Piek Vossen. 2014.
\newblock \href
  {http://www.lrec-conf.org/proceedings/lrec2014/pdf/840_Paper.pdf} {Using a
  sledgehammer to crack a nut? lexical diversity and event coreference
  resolution}.
\newblock In \emph{Proceedings of the Ninth International Conference on
  Language Resources and Evaluation ({LREC}-2014)}, pages 4545--4552,
  Reykjavik, Iceland. European Languages Resources Association (ELRA).

\bibitem[{Cybulska and Vossen(2015)}]{cybulska-vossen-2015-translating}
Agata Cybulska and Piek Vossen. 2015.
\newblock \href {https://doi.org/10.3115/v1/W15-0801} {Translating granularity
  of event slots into features for event coreference resolution.}
\newblock In \emph{Proceedings of the The 3rd Workshop on {EVENTS}: Definition,
  Detection, Coreference, and Representation}, pages 1--10, Denver, Colorado.
  Association for Computational Linguistics.

\bibitem[{Devlin et~al.(2019)Devlin, Chang, Lee, and
  Toutanova}]{devlin-etal-2019-bert}
Jacob Devlin, Ming-Wei Chang, Kenton Lee, and Kristina Toutanova. 2019.
\newblock \href {https://doi.org/10.18653/v1/N19-1423} {{BERT}: Pre-training of
  deep bidirectional transformers for language understanding}.
\newblock In \emph{Proceedings of the 2019 Conference of the North {A}merican
  Chapter of the Association for Computational Linguistics: Human Language
  Technologies, Volume 1 (Long and Short Papers)}, pages 4171--4186,
  Minneapolis, Minnesota. Association for Computational Linguistics.

\bibitem[{Dhingra et~al.(2018)Dhingra, Jin, Yang, Cohen, and
  Salakhutdinov}]{dhingra-etal-2018-neural}
Bhuwan Dhingra, Qiao Jin, Zhilin Yang, William Cohen, and Ruslan Salakhutdinov.
  2018.
\newblock \href {https://doi.org/10.18653/v1/N18-2007} {Neural models for
  reasoning over multiple mentions using coreference}.
\newblock In \emph{Proceedings of the 2018 Conference of the North {A}merican
  Chapter of the Association for Computational Linguistics: Human Language
  Technologies, Volume 2 (Short Papers)}, pages 42--48, New Orleans, Louisiana.
  Association for Computational Linguistics.

\bibitem[{Falke et~al.(2017)Falke, Meyer, and
  Gurevych}]{falke-etal-2017-concept}
Tobias Falke, Christian~M. Meyer, and Iryna Gurevych. 2017.
\newblock \href {https://www.aclweb.org/anthology/I17-1081} {Concept-map-based
  multi-document summarization using concept coreference resolution and global
  importance optimization}.
\newblock In \emph{Proceedings of the Eighth International Joint Conference on
  Natural Language Processing (Volume 1: Long Papers)}, pages 801--811, Taipei,
  Taiwan. Asian Federation of Natural Language Processing.

\bibitem[{Glorot and Bengio(2010)}]{glorot2010understanding}
Xavier Glorot and Yoshua Bengio. 2010.
\newblock Understanding the difficulty of training deep feedforward neural
  networks.
\newblock In \emph{Proceedings of the thirteenth international conference on
  artificial intelligence and statistics}, pages 249--256.

\bibitem[{Joshi et~al.(2020)Joshi, Chen, Liu, Weld, Zettlemoyer, and
  Levy}]{doi:10.1162/tacl_a_00300}
Mandar Joshi, Danqi Chen, Yinhan Liu, Daniel~S. Weld, Luke Zettlemoyer, and
  Omer Levy. 2020.
\newblock Spanbert: Improving pre-training by representing and predicting
  spans.
\newblock \emph{Transactions of the Association for Computational Linguistics},
  8:64--77.

\bibitem[{Joshi et~al.(2019)Joshi, Levy, Zettlemoyer, and
  Weld}]{joshi-etal-2019-bert}
Mandar Joshi, Omer Levy, Luke Zettlemoyer, and Daniel Weld. 2019.
\newblock \href {https://doi.org/10.18653/v1/D19-1588} {{BERT} for coreference
  resolution: Baselines and analysis}.
\newblock In \emph{Proceedings of the 2019 Conference on Empirical Methods in
  Natural Language Processing and the 9th International Joint Conference on
  Natural Language Processing (EMNLP-IJCNLP)}, pages 5803--5808, Hong Kong,
  China. Association for Computational Linguistics.

\bibitem[{Kantor and Globerson(2019)}]{kantor-globerson-2019-coreference}
Ben Kantor and Amir Globerson. 2019.
\newblock \href {https://doi.org/10.18653/v1/P19-1066} {Coreference resolution
  with entity equalization}.
\newblock In \emph{Proceedings of the 57th Annual Meeting of the Association
  for Computational Linguistics}, pages 673--677, Florence, Italy. Association
  for Computational Linguistics.

\bibitem[{Kenyon-Dean et~al.(2018)Kenyon-Dean, Cheung, and
  Precup}]{kenyon-dean-etal-2018-resolving}
Kian Kenyon-Dean, Jackie Chi~Kit Cheung, and Doina Precup. 2018.
\newblock \href {https://doi.org/10.18653/v1/S18-2001} {Resolving event
  coreference with supervised representation learning and clustering-oriented
  regularization}.
\newblock In \emph{Proceedings of the Seventh Joint Conference on Lexical and
  Computational Semantics}, pages 1--10, New Orleans, Louisiana. Association
  for Computational Linguistics.

\bibitem[{Kingma and Ba(2014)}]{kingma2014adam}
Diederik~P Kingma and Jimmy Ba. 2014.
\newblock Adam: A method for stochastic optimization.
\newblock \emph{arXiv preprint arXiv:1412.6980}.

\bibitem[{Lee et~al.(2012)Lee, Recasens, Chang, Surdeanu, and
  Jurafsky}]{lee-etal-2012-joint}
Heeyoung Lee, Marta Recasens, Angel Chang, Mihai Surdeanu, and Dan Jurafsky.
  2012.
\newblock \href {https://www.aclweb.org/anthology/D12-1045} {Joint entity and
  event coreference resolution across documents}.
\newblock In \emph{Proceedings of the 2012 Joint Conference on Empirical
  Methods in Natural Language Processing and Computational Natural Language
  Learning}, pages 489--500, Jeju Island, Korea. Association for Computational
  Linguistics.

\bibitem[{Lee et~al.(2017)Lee, He, Lewis, and Zettlemoyer}]{lee-etal-2017-end}
Kenton Lee, Luheng He, Mike Lewis, and Luke Zettlemoyer. 2017.
\newblock \href {https://doi.org/10.18653/v1/D17-1018} {End-to-end neural
  coreference resolution}.
\newblock In \emph{Proceedings of the 2017 Conference on Empirical Methods in
  Natural Language Processing}, pages 188--197, Copenhagen, Denmark.
  Association for Computational Linguistics.

\bibitem[{Lee et~al.(2018)Lee, He, and Zettlemoyer}]{lee2018higher}
Kenton Lee, Luheng He, and Luke~S. Zettlemoyer. 2018.
\newblock Higher-order coreference resolution with coarse-to-fine inference.
\newblock In \emph{Proceedings of the 2018 Annual Conference of the North
  American Chapter of the Association for Computational Linguistics}.

\bibitem[{Liu et~al.(2019)Liu, Ott, Goyal, Du, Joshi, Chen, Levy, Lewis,
  Zettlemoyer, and Stoyanov}]{liu2019roberta}
Yinhan Liu, Myle Ott, Naman Goyal, Jingfei Du, Mandar Joshi, Danqi Chen, Omer
  Levy, Mike Lewis, Luke Zettlemoyer, and Veselin Stoyanov. 2019.
\newblock Roberta: A robustly optimized bert pretraining approach.
\newblock \emph{arXiv preprint arXiv:1907.11692}.

\bibitem[{Luo(2005)}]{luo-2005-coreference}
Xiaoqiang Luo. 2005.
\newblock \href {https://www.aclweb.org/anthology/H05-1004} {On coreference
  resolution performance metrics}.
\newblock In \emph{Proceedings of Human Language Technology Conference and
  Conference on Empirical Methods in Natural Language Processing}, pages
  25--32, Vancouver, British Columbia, Canada. Association for Computational
  Linguistics.

\bibitem[{Meged et~al.(2020)Meged, Caciularu, Shwartz, and
  Dagan}]{Meged2020ParaphrasingVC}
Y.~Meged, Avi Caciularu, Vered Shwartz, and I.~Dagan. 2020.
\newblock Paraphrasing vs coreferring: Two sides of the same coin.
\newblock \emph{ArXiv}, abs/2004.14979.

\bibitem[{Moosavi and Strube(2016)}]{moosavi-strube-2016-coreference}
Nafise~Sadat Moosavi and Michael Strube. 2016.
\newblock \href {https://doi.org/10.18653/v1/P16-1060} {Which coreference
  evaluation metric do you trust? a proposal for a link-based entity aware
  metric}.
\newblock In \emph{Proceedings of the 54th Annual Meeting of the Association
  for Computational Linguistics (Volume 1: Long Papers)}, pages 632--642,
  Berlin, Germany. Association for Computational Linguistics.

\bibitem[{Paszke et~al.(2017)Paszke, Gross, Chintala, Chanan, Yang, DeVito,
  Lin, Desmaison, Antiga, and Lerer}]{paszke2017automatic}
Adam Paszke, Sam Gross, Soumith Chintala, Gregory Chanan, Edward Yang, Zachary
  DeVito, Zeming Lin, Alban Desmaison, Luca Antiga, and Adam Lerer. 2017.
\newblock Automatic differentiation in pytorch.

\bibitem[{Pradhan et~al.(2012)Pradhan, Moschitti, Xue, Uryupina, and
  Zhang}]{pradhan-etal-2012-conll}
Sameer Pradhan, Alessandro Moschitti, Nianwen Xue, Olga Uryupina, and Yuchen
  Zhang. 2012.
\newblock \href {https://www.aclweb.org/anthology/W12-4501} {{C}o{NLL}-2012
  shared task: Modeling multilingual unrestricted coreference in
  {O}nto{N}otes}.
\newblock In \emph{Joint Conference on {EMNLP} and {C}o{NLL} - Shared Task},
  pages 1--40, Jeju Island, Korea. Association for Computational Linguistics.

\bibitem[{Shwartz et~al.(2017)Shwartz, Stanovsky, and
  Dagan}]{shwartz-etal-2017-acquiring}
Vered Shwartz, Gabriel Stanovsky, and Ido Dagan. 2017.
\newblock \href {https://doi.org/10.18653/v1/S17-1019} {Acquiring predicate
  paraphrases from news tweets}.
\newblock In \emph{Proceedings of the 6th Joint Conference on Lexical and
  Computational Semantics (*{SEM} 2017)}, pages 155--160, Vancouver, Canada.
  Association for Computational Linguistics.

\bibitem[{Upadhyay et~al.(2016)Upadhyay, Gupta, Christodoulopoulos, and
  Roth}]{upadhyay-etal-2016-revisiting}
Shyam Upadhyay, Nitish Gupta, Christos Christodoulopoulos, and Dan Roth. 2016.
\newblock \href {https://www.aclweb.org/anthology/C16-1183} {Revisiting the
  evaluation for cross document event coreference}.
\newblock In \emph{Proceedings of {COLING} 2016, the 26th International
  Conference on Computational Linguistics: Technical Papers}, pages 1949--1958,
  Osaka, Japan. The COLING 2016 Organizing Committee.

\bibitem[{Wang et~al.(2019)Wang, Yu, Guo, Das, Xiong, and
  Gao}]{wang-etal-2019-multi-hop}
Haoyu Wang, Mo~Yu, Xiaoxiao Guo, Rajarshi Das, Wenhan Xiong, and Tian Gao.
  2019.
\newblock \href {https://doi.org/10.18653/v1/D19-5813} {Do multi-hop readers
  dream of reasoning chains?}
\newblock In \emph{Proceedings of the 2nd Workshop on Machine Reading for
  Question Answering}, pages 91--97, Hong Kong, China. Association for
  Computational Linguistics.

\bibitem[{Wolf et~al.(2019)Wolf, Debut, Sanh, Chaumond, Delangue, Moi, Cistac,
  Rault, Louf, Funtowicz et~al.}]{wolf2019transformers}
Thomas Wolf, Lysandre Debut, Victor Sanh, Julien Chaumond, Clement Delangue,
  Anthony Moi, Pierric Cistac, Tim Rault, R{\'e}mi Louf, Morgan Funtowicz,
  et~al. 2019.
\newblock Transformers: State-of-the-art natural language processing.
\newblock \emph{arXiv preprint arXiv:1910.03771}.

\bibitem[{Wu et~al.(2020)Wu, Wang, Yuan, Wu, and Li}]{wu-etal-2020-corefqa}
Wei Wu, Fei Wang, Arianna Yuan, Fei Wu, and Jiwei Li. 2020.
\newblock \href {https://doi.org/10.18653/v1/2020.acl-main.622} {{C}oref{QA}:
  Coreference resolution as query-based span prediction}.
\newblock In \emph{Proceedings of the 58th Annual Meeting of the Association
  for Computational Linguistics}, pages 6953--6963, Online. Association for
  Computational Linguistics.

\bibitem[{Yang et~al.(2015)Yang, Cardie, and
  Frazier}]{yang-etal-2015-hierarchical}
Bishan Yang, Claire Cardie, and Peter Frazier. 2015.
\newblock \href {https://doi.org/10.1162/tacl_a_00155} {A hierarchical
  distance-dependent {B}ayesian model for event coreference resolution}.
\newblock \emph{Transactions of the Association for Computational Linguistics},
  3:517--528.

\end{thebibliography}

\appendix

\section{Dataset}
\label{app:dataset}

We follow previous work and use the ECB+ corpus for our experiments, statistics are shown in Table~\ref{tab:ecb_stat}. ECB+ is publicly available\footnote{\url{http://www.newsreader-project.eu/results/data/the-ecb-corpus/}} and was built upon the Event Coreference Bank~(ECB;~\citealp{bejan-harabagiu-2008-linguistic}) and the Extended ECB~(EECB;~\citealp{lee-etal-2012-joint}). 

As opposed to OntoNotes, only a few sentences are exhaustively annotated in each document, and the annotations include singletons. Also, entities are only annotated if they participate in an event in the annotated sentence (event participants)

\begin{table}[!h]
    \centering
    \resizebox{0.48\textwidth}{!}{
    \begin{tabular}{@{}llll@{}}
    \toprule
    & \textbf{Train} & \textbf{Validation} & \textbf{Test} \\
    \midrule
    \# Topics & 25 & 8 & 10 \\
    \# Documents & 594 & 196 & 206 \\
    \# Sentences & 1037 & 346 & 457 \\
    \# Mentions & 3808/4758 & 1245/1476 & 1780/2055 \\
    \# Singletons & 1116/814 & 280/205 & 632/412 \\
    \# Clusters & 411/472 & 129/125 & 182/196 \\
    \bottomrule
    \end{tabular}}
    \caption{ECB+ statistics. \# Clusters do not include singletons. The slash numbers for \# Mentions, \# Singletons, and \# Clusters represent event/entity statistics. As recommended by the authors in the release note, we follow the split of \citet{cybulska-vossen-2015-translating} that use a curated subset of the dataset.}
    \label{tab:ecb_stat}
\end{table}

\section{Singleton Effect}
\label{app:singleton}

    


\begin{table*}[!ht]
    \centering
    \resizebox{\textwidth}{!}{
    \begin{tabular}{@{}llllllllllllllllll@{}}
    \toprule

    
     \phantom{abcdef} &&& \multicolumn{3}{c}{MUC} && \multicolumn{3}{c}{$B^3$} && \multicolumn{3}{c}{$CEAFe$} && \multicolumn{3}{c}{LEA}\\
    
    \cmidrule{4-6} \cmidrule{8-10} \cmidrule{12-14} \cmidrule{16-18}
    
    &&& R & P & $F_1$ && R & P & $F_1$ && R &P & $F_1$ && R & P & $F_1$  \\ 
    
    \midrule 
        \multirow{2}{*}{With Singletons} & S1 && 100 & 60.0 & 75.0 && 100 & 63.3 & \textbf{77.6} && 66.7 & 93.3 & \textbf{77.8} && 90.0 & 56.0 & \textbf{69.0} \\
        & S2 && 100 & 75.0 & \textbf{85.7} && 60.0 & 66.7 & 63.2 && 25.7 & 60.0 & 36.0 && 50.0 & 57.1 & 53.3 \\
        \midrule
        \multirow{2}{*}{Without Singletons} & S1 && 100 & 60.0 & 75.0 && 100 & 36.1 & 53.1 && 33.3 & 66.7 & 44.4 && 100 & 26.7 & 42.1 \\
        & S2 && 100 & 75.0 & \textbf{85.7} && 100 & 72.2 & \textbf{83.9} && 90.0 & 90.0 & \textbf{90.0} && 100 & 66.7 & \textbf{80.0} \\
    \bottomrule
    \end{tabular}}
    \caption{Coreference results of S1 and S2 with and without singletons.}
    \label{tab:singletons}
\end{table*}

We show the \emph{singleton effect} by creating two hypothetical predictions by two different systems $S1$ and $S2$ on a realistic setting. $S1$ is good at making local decisions such as named entity spans, but bad at linking coreferring mentions, while $S2$ performs the coreference task better, but struggles at predicting singletons. 
Specifically, assume the following example of gold clusters \{\{A\}, \{B\}, \{C\}, \{D\}, \{E\}, \{F, G\}, \{H, I, J\}\}\footnote{This example follows the distribution of singletons in natural texts (about 50\%), as illustrated in the PreCo dataset~\citep{chen-etal-2018-preco}} and the output of two systems S1 and S2: \\\\
$S1$: $\{A\}, \{B\}, \{C\}, \{D\}, \{E, F, G, H, I, J\}$\\
$S2$: $\{E, F, G\}, \{H, I, J\}, \{Z\}$\\

With respect to singletons, $S1$ identified the exact spans of the singleton clusters $\{A\}, \{B\}, \{C\}, \{D\}$, while $S2$ missed them and predicted a wrong span ($Z$) as singleton mention.
Both $S1$ and $S2$ have erroneously merged the singleton mention E with the cluster $\{F, G\}$, however, $S1$ has further mixed these mentions with the cluster $\{H, I, J\}$, whereas $S2$ successfully predicted $\{H, I, J\}$ in their own cluster. 
As explained in the paper (Section~\ref{subsec:wd-standard}), singleton prediction does not have any downstream impact, and since $S2$ performs better on \emph{linking} the mentions, it should be considered a better model.

Table~\ref{tab:singletons} shows the results of $S1$ and $S2$ when including or omitting singletons in the evaluation. Apart from MUC (link-based metric), the scores w.r.t B\textsuperscript{3}, CEAF-e, and LEA differ significantly when including or omitting singletons for both $S1$ and $S2$. More importantly, the performance of $S1$ when including singletons, is higher than $S2$ when including singletons, which is counterproductive with the aforementioned downstream goal of coreference resolution. This phenomenon stems from the large proportion of singletons, where each one is rewarded 100\% in both recall and precision. 
In contrast, when excluding singletons, the expected behavior is achieved, $S2$ gets the best performance, and the singleton baseline achieves 0. 
It is worth noting that even when excluding singletons, models are still penalized for making coreference errors involving singletons (as $S2$ is penalized from linking $E$ to a cluster). 

As observed by~\citet{moosavi-strube-2016-coreference}, the mention-based evaluation metrics B\textsuperscript{3} and CEAFe suffer also from the \emph{mention identification effect}---rewarding a coreference model because a predicted mention exists also in the gold regardless of whether it has a correct coreference relation. While the \emph{mention identification effect} does not apply to their new evaluation metric, LEA, singletons should be removed before the evaluation to address the \emph{singleton effect}. 

Finally, we note that while singletons distorts the coreference evaluation, their annotation is still valuable for the \emph{mention detection} task, be either entity or event, when trained and evaluated separately. In this case, $S1$ will obviously get higher scores than $S2$.

\section{Experimental setting}
\label{app:model}

Our model includes 14M parameters and is implemented in PyTorch \cite{paszke2017automatic}, using HuggingFace's library~\cite{wolf2019transformers} and the Adam optimizer \cite{kingma2014adam}. 
The layers of the models are initialized with Xavier Glorot method \cite{glorot2010understanding}.  
We manually tuned the standard hyperparameters, presented in Table~\ref{tab:shared} on the event coreference task and keep them unchanged for entity and ALL settings.
Table~\ref{tab:specific} shows specific parameters, such as the maximum span width, the pruning coefficient $\lambda$ and the stop criterion $\tau$ for the agglomerative clustering, that we tuned separately for each setting to maximize the CoNLL F1 score on its corresponding development set.

\begin{table}[!]
    \centering
    \begin{tabular}{lll}
    \toprule
    Hyperparameter   &\phantom{abckjbn}  &  Value\\
    \midrule
    Batch size && 32 \\
    Dropout && 0.3 \\
    Learning rate && 0.001\\
    Optimizer && Adam \\
    Hidden layer && 1024 \\
    \bottomrule
    \end{tabular}
    \caption{Shared hyperparameters across the different models. }
    \label{tab:shared}
\end{table}

\begin{table}[!]
    \centering
    \begin{tabular}{lllllll}
    \toprule
    && Max span width && $\lambda$ && $\tau$ \\
    \midrule
    Event  &&  10 && 0.25 && 0.65 \\
    Entity  &&  15 && 0.35 && 0.6 \\
    ALL  &&  15 && 0.4 && 0.55 \\
    \bottomrule
    \end{tabular}
    \caption{Specific hyperparameters for each mention type; $\lambda$ is the pruning coefficient and $\tau$ is the threshold for the agglomerative clustering. }
    \label{tab:specific}
\end{table}

\end{document}